\pdfoutput=1

\documentclass[11pt]{article}

\usepackage{acl}

\usepackage{booktabs}
\usepackage{times}
\usepackage{latexsym}
\usepackage{enumitem}
\usepackage[hang,flushmargin]{footmisc}
\usepackage{wrapfig}
\usepackage[framemethod=tikz]{mdframed}
\usepackage{verbatimbox}
\usepackage{hyperref}
\usepackage[T1]{fontenc}

\usepackage[utf8]{inputenc}

\usepackage{microtype}

\usepackage{inconsolata}

\usepackage[utf8]{inputenc} 
\usepackage[T1]{fontenc}    
\usepackage{hyperref}       
\usepackage{url}            
\usepackage{booktabs}       
\usepackage{amsfonts}       
\usepackage{amsmath}
\usepackage{nicefrac}       
\usepackage{microtype}      
\usepackage{xcolor}         
\usepackage{amssymb}
\usepackage{graphics}
\usepackage{listings}
\usepackage{colortbl}
\usepackage{csquotes}
\usepackage{cleveref}
\usepackage{tikz}
\usepackage{float}
\usepackage{pifont}
\usepackage{array}
\usepackage{multirow}
\usepackage{xspace}
\newcolumntype{L}{>{\arraybackslash}m{13cm}}

\usepackage{booktabs}
\usepackage{siunitx}
\usepackage{amssymb}    
\usepackage{tikz}
\usepackage{pdfpages}
\usetikzlibrary{shapes.geometric, arrows, positioning}
\newcommand{\cmark}{\ding{51}}%
\newcommand{\xmark}{\ding{55}}%
\newcommand{\convkg}{Conv\-KG\-Yarn\xspace}
\newcommand{\convkgg}{Conv\-KG\-Yarn\textsubscript{G}\xspace}
\newcommand{\convkgr}{Conv\-KG\-Yarn\textsubscript{R}\xspace}

\newcommand{\convquestions}{Conv\-Questions\xspace}

\newcommand{\voice}{Voice\xspace}
\newcommand{\search}{Text\xspace}

\newcommand{\gptfour}{GPT\textsubscript{4}\xspace}
\newcommand{\gptthreefive}{GPT\textsubscript{3.5}\xspace}

\definecolor{anti-flashwhite}{rgb}{0.95, 0.95, 0.96}
\definecolor{lightred}{rgb}{1.0, 0.25, 0.25}

\mdfdefinestyle{MyFrame}{%
    outerlinewidth=2pt,
    roundcorner=10pt,
    linecolor=blue,
    innertopmargin=10pt,
    innerbottommargin=10pt,
    innerrightmargin=10pt,
    innerleftmargin=10pt,
    backgroundcolor=gray!20!white}

\usepackage[framemethod=tikz]{mdframed}
\usepackage{listings} 

\lstnewenvironment{wrappedverbatim}[1][]
{
  \lstset{basicstyle=\ttfamily,columns=fullflexible,keepspaces=true,#1}
}
{}

\title{ConvKGYarn: Spinning Configurable and Scalable Conversational Knowledge Graph QA datasets with Large Language Models}

\author{Ronak Pradeep$^{1,2}$, Daniel Lee$^{1,3}$, Ali Mousavi$^{1}$, Jeff Pound$^{1}$, Yisi Sang$^{1}$,  \\
{\bf Jimmy Lin$^{2}$, Ihab Ilyas$^{1}$, Saloni Potdar$^{1}$, Mostafa Arefiyan$^{1}$} \and {\bf Yunyao Li$^{4}$\thanks{Work done while at Apple}} \\[0.5ex]
  $^1$ Apple \quad $^2$ University of Waterloo \quad $^3$ University of Calgary \quad $^4$ Adobe \\[0.5ex]
  \texttt{rpradeep@uwaterloo.ca} \texttt{\{mostafaa, s\_potdar\}@apple.com} \\ \texttt{yunyaol@adobe.com}}
\begin{document}
\maketitle

\begin{abstract}
The rapid advancement of Large Language Models (LLMs) and conversational assistants necessitates dynamic, scalable, and configurable conversational datasets for training and evaluation.
These datasets must accommodate diverse user interaction modes, including text and voice, each presenting unique modeling challenges. 
Knowledge Graphs (KGs), with their structured and evolving nature, offer an ideal foundation for current and precise knowledge.
Although human-curated KG-based conversational datasets exist, they struggle to keep pace with the rapidly changing user information needs.
We present \convkg, a scalable method for generating up-to-date and configurable conversational KGQA datasets. 
Qualitative psychometric analyses confirm our method can generate high-quality datasets rivaling a popular conversational KGQA dataset while offering it at scale and covering a wide range of human-interaction configurations.
We showcase its utility by testing LLMs on diverse conversations --- exploring model behavior on conversational KGQA sets with different configurations grounded in the same KG fact set.
Our results highlight the ability of \convkg to improve KGQA foundations and evaluate parametric knowledge of LLMs, thus offering a robust solution to the constantly evolving landscape of conversational assistants.
\end{abstract}

\section{Introduction}
\label{sec:introduction}

The proliferation of Large Language Models (LLMs) and conversational assistants has led to their ubiquitous presence in daily user interactions. This widespread adoption underscores the critical need for dynamic datasets capable of rigorously evaluating their proficiency in addressing knowledge-seeking queries.
Knowledge Graphs (KGs) have long been recognized as powerful tools for capturing structured representations of the world~\cite{hogan-etal-2021-knowledge}.
In KGs, concepts and entities are represented as nodes, while semantic relationships defining facts are represented with edges. 
This structured representation has had an impact across various domains, including Natural Language Processing~\cite{schneider-etal-2022-decade}, Recommender Systems~\cite{guo-etal-2022-recommender}, and Information Retrieval~\cite{reinanda-etal-2020-knowledge}.

\begin{figure*}[t]
    \centering
    \includegraphics[width=0.95\linewidth]{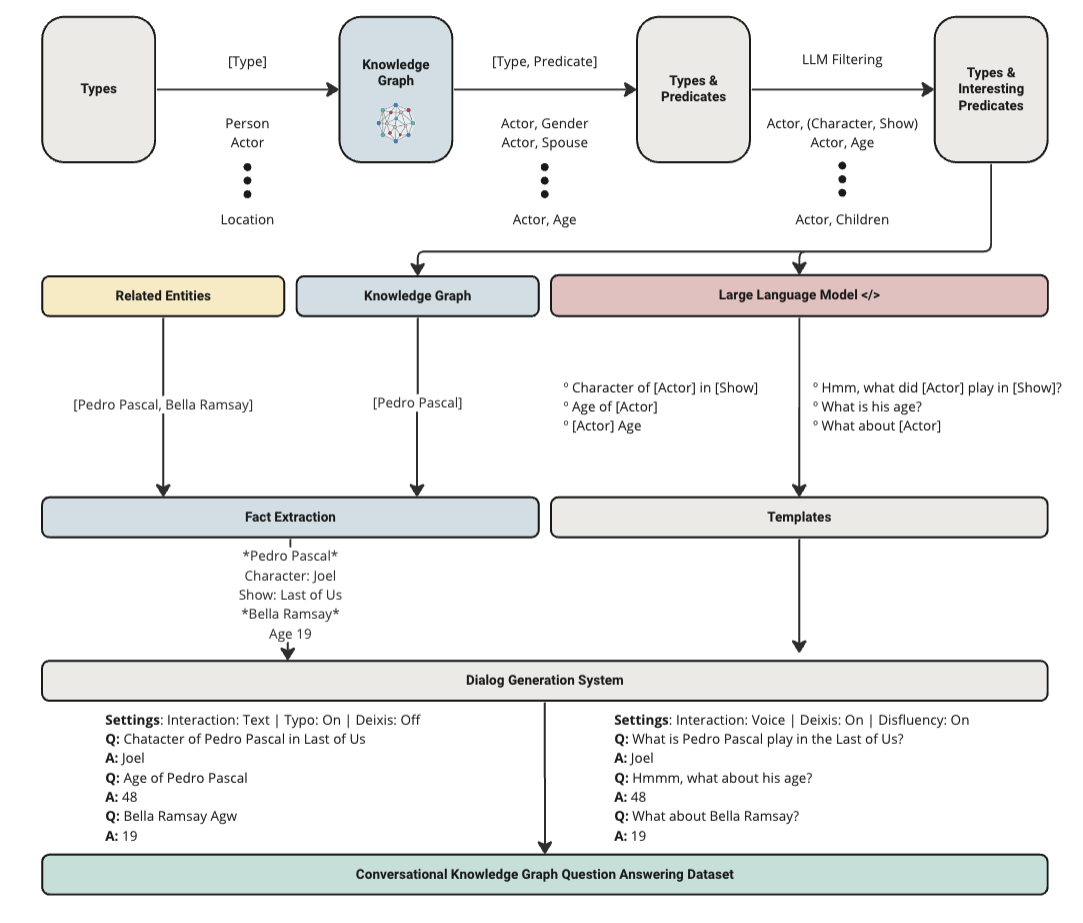}
    \caption{The full \convkg pipeline.}
    \label{fig:pipeline}
\end{figure*}
The integration of LLMs and KGs has opened up new opportunities in natural language processing~\cite{petroni-etal-2019-language,guu-etal-2020-realm,peng-etal-2023-check}, which has led to significant advancements across various tasks~\cite{barba-etal-2021-esc,chakrabarti-etal-2022-joint,de-cao-etal-2022-multilingual,xu-etal-2023-kilm}. 
By combining the dynamic capabilities of LLMs with the structured insights from KGs, researchers have unlocked new avenues for developing advanced question-answering (QA) systems.
In conversational Knowledge Graph Question Answering (KGQA), datasets like ConvQuestions~\cite{10.1145/3357384.3358016} have emerged to address scenarios where questions often lack full context or contain grammatical inconsistencies. 
These datasets have played a crucial role in enabling new retrieval-augmented systems, demonstrating the potential of LLM-KG integrations to provide accurate and attributable responses in conversational settings~\cite{christmann2023compmix}.

In relation, advancements in text retrieval have underscored the potential of using LLMs to generate synthetic data to improve the effectiveness of downstream systems.
This process has been utilized at scale for neural query synthesis~\cite{doct5query, unicoil, NQS} and LLM-based ranked list reorderings for instruction distillation into open-source rerankers~\cite{rankvicuna, rankzephyr, lit5} resulting in substantial improvements across a spectrum of retrieval tasks. 
More recently, synthetic data generation, facilitated by automated prompt optimization~\cite{path}, has enabled the training of highly effective small-scale models, deprecating the dependence on human-labeled data. 
Collectively, these highlight the opportunity to integrate synthetic data strategies from LLMs to develop more resilient and adaptable knowledge-intensive systems.

While existing conversational KGQA datasets are rich in content, they often struggle to keep pace with rapidly evolving user information needs. 
This discrepancy raises questions about the relevance of such data in real-world, adaptive conversational scenarios.
To address this challenge, we introduce \convkg, a novel method for generating large-scale, configurable conversational Knowledge Graph Question Answering (KGQA) datasets.
Through rigorous psychometric evaluation metrics, we demonstrate that \convkg produces high-quality conversational data comparable to established, human-curated KGQA datasets. 
Notably, \convkg achieves this while significantly expanding entity and fact coverage by several orders of magnitude and introducing configurable properties in user interaction styles.

A critical component of our research involves evaluating the datasets generated by \convkg using various LLMs to assess their parametric knowledge. 
Our observations reveal that these models often struggle with fact recall, underscoring the need for retrieval-augmented systems.
By evaluating datasets generated with different interaction styles and their varied linguistic phenomena generated with consistent fact sets from the Knowledge Graph, we aim to assess the robustness of LLMs in handling diverse conversational settings, in confounder-free manners.
Our findings indicate that model effectiveness varies significantly across diverse user interaction styles, highlighting the importance of developing LLMs capable of serving as robust conversational systems.

Through this work, we seek to reveal the path toward creating datasets that can effectively train and evaluate evolving conversational assistants.
We envision these methods and datasets will play a crucial role in developing more versatile and adaptive conversational AI systems.
\section{\convkg}
\label{sec:method}

\autoref{fig:pipeline} illustrates the entire pipeline of the \convkg system. 
We first introduce the key notations and definitions that are of use in our framework.
Following this, we dive into each module that comprises the \convkg system.
\subsection{Definitions and Notations}
The knowledge graph (KG) acts as the foundation for our work.
Following the Wikidata terminology, we have an item (or entity), \( e \in \mathcal{E} \) and statements (or facts) \( \mathcal{S}_{e} \), that we represent by \emph{item-property-value} tuples that help describe the fact.
The properties (or predicates) are denoted by \( p\textsubscript{e} \in \mathcal{P}\textsubscript{e} \).

We denote values (or objects) for a particular entity and predicate $p_e$, with $o_{p_e}$.
In \convkg, we use a \emph{simple fact} to denote item-property-value tuples where the corresponding predicate does not involve multiple entries. 
Conversely, some entities naturally possess properties with multiple values, such as the siblings of an Actor or the official languages of a Country. 
These are acceptable within \convkg as \emph{complex facts}. 
Another exception, which we dub \emph{qualified facts}, is where different values hold in light of qualifiers (for example, the population of a Country or the CEO of a company can both include qualifiers by timestamps).
These qualifiers further delineate or refine the values within a statement and are accepted in the \convkg framework.

Each entity \( e \) is associated with multiple \emph{types} \( T_e \) with a specific type denoted by \( t_{e} \).
Examples of types are Person, Singer, Book, and Movie.
Note that in \convkg, in addition to using the \texttt{InstanceOf} predicate to describe types, we also leverage the  \texttt{Occupation} predicate to add nuances to the types, and from there, the predicates we draw our attention to since we believe that the \emph{interesting} predicates for someone who is a Politician are likely different from one who is an Actor. 
\subsection{KG Predicate Extraction}
The initial stage of \convkg leverages the KG to extract all predicates \( p_i \) for a particular entity type \( T \). 
This extraction process is denoted by \( \mathcal{F}(t) = \{ p_1, p_2, \ldots, p_n \} \), where \( \mathcal{F} \) is the extraction function, \( t \) is a type, and \( \{ p_1, p_2, \ldots, p_n \} \) is the set of predicates such that there exists some entity \( e \) of type \( t \), for which $p_i$ is a valid predicate.
In our example in \autoref{fig:pipeline}, we see that for an Actor, these include predicates like Gender or Spouse.
\subsection{LLM Predicate Selector}
This step employs a large language model (LLM) to filter the extracted predicates, selecting only the most \emph{interesting} predicates for each entity type. 
This process is governed by the prompt detailed in \autoref{fig:prompt.pfilter} in \autoref{sec:add_prompts} and can be formulated as \( \mathcal{G}(t, \{ p_1, p_2, \ldots, p_n \}) = \{ p'_1, p'_2, \ldots, p'_m \} \),
where \( \mathcal{G} \) represents the selector function, selecting a subset \( \{ p'_1, p'_2, \ldots, p'_m \} \) from the initial set of predicates.

By prompting with a high degree of specificity, we hope the selection optimizes for predicates that enhance the richness of the dataset while maintaining contextual appropriateness.
Additionally, by including the Wikidata identifier of the predicate, we hope to resolve cases where the identifier name is unclear, especially given that these models have most likely encountered them during training.

Furthermore, we expect the predicates that pass through this filter to contribute meaningfully to conversations surrounding the entity type.
To achieve this, we prompt the model to exclude predicates that are too generic, irrelevant noise, or identifiers, none of which lend themselves to a high-quality conversational QA dataset in any interaction form.

We see in \autoref{fig:pipeline} that for an Actor, \convkg selects predicates like (Character, Show), a qualified predicate, or Age.

\subsection{Related Entity Generator}
The related entity generator \( \mathcal{R} \) is an additional component of \convkg that identifies and selects entities \( e_r \) linked to the primary entity \( e \). 
Doing this allows for the enrichment of the dataset with diverse but relevant information that is often not directly in the vicinity of the original entity (for example, as seen in \autoref{fig:pipeline}, actors like Pedro Pascal and Bella Ramsay might not be direct neighbors on Wikidata graph, yet questions about them could show up in the same conversation by their association through the Last of Us TV series).
Related entities can be selected using KG embedding similarity (inner product) with embeddings that prioritize capturing the ontology of the graph. 
We use only the \emph{most-similar} related entity for popular Person entities to not introduce bias or excessive noise into our datasets. 
\subsection{Fact Extraction}
Using the KG, \convkg extracts factual information \( \mathcal{I} \) corresponding to each entity.
For an entity \( e \), we represent the fact extraction for simple or complex facts by \( \mathcal{I}(e) = \{ (e, p'_1, o_1), \ldots, (e, p'_m, o_m) \}\),
where \( o_i \) denotes the object(s) corresponding to the ``interesting'' predicate \( p'_i \). 

In the case of \emph{qualified facts}, we can generalize this to include \(\mathcal{I}_{c}(e) = \bigcup_{i=1}^m \bigcup_{j=1}^{l_i} \{(e, p'_i, q_{i}, o_{i})\} \),
where $q_{i}$ is the qualifier set.
\subsection{Synthetic Question Template Generation}
To maintain configurability and scalability, in addition to ensuring the tractability of \convkg framework, we generate questions using a templated approach that incorporates placeholder entity type (with actual type information in natural language form, for example, [actor]), interesting predicates, and placeholder objects ([i]) in the prompt.
For simple and complex facts, the detailed prompt for generating questions for voice interactions and textual (or search) interactions are in \autoref{fig:prompt.question} and \autoref{fig:prompt.query} in \autoref{sec:add_prompts}, respectively. 
The prompt for qualified facts is in \autoref{fig:prompt.complex_query} (in Appendix~\ref{sec:add_prompts}).

In designing \convkg, it was imperative to emulate the nuances of both text and voice interactions, representing the primary modalities through which users engage with AI assistants.
    The goal was to capture the essence of these interactions with the prompts, spotlighting the differences in user experience. 
   In text interactions, we aim to mimic search queries, emphasizing short keyword queries and with follow-ups made in succession.
    These interactions allow for \emph{deixis}, where questions reflect references to previously mentioned entities in the conversation, enhancing their continuity.
    Additionally, \convkg also has a knob for typographical errors (typos), another phenomenon common in textual interactions, albeit in a post-processing step discussed in Section~\ref{sec:exp}.
   In voice interactions, we hope the system generates more well-formed questions. 
   In addition to \emph{deixis}, the voice modality allows for conversations with \emph{disfluencies}.
    Disfluencies aim to mimic the imperfections of natural speech by adding natural uses of \emph{uh}, \emph{um}, takebacks, apologies, thanks, or repetitions, among others. 
    We hope to amalgamate these aspects in the ``deixis\_disfluencies'' variants to simulate the intricacies of human conversation, involving both references and speech errors.
    
    The structured prompt ensures that for each fact and linguistic phenomenon, we generate three question variants. 
    Doing so ensures more variations in the generated questions versus sampling questions by querying the LLM multiple times, which is slower, more expensive, and less guaranteed to output variants.
    Additionally, by generating all variants together, we hope to have variants with linguistic phenomena that build on the same original variant.
    This approach helps better evaluate the robustness of conversational QA systems or LLMs while providing comprehensive training data that includes a wide range of linguistic variations.

   We speed up inference by providing five triples instead to save on the bulk of the input tokens (the instructions).
Note that the turn number does not mean \convkg is generating a question for that specific turn, although they do capture sequential order important for \emph{qualified facts}.
    Instead, its purpose is to give an index for both the JSON key and the object identifier.
    
    The JSON format used in the prompt is pivotal for systematic data parsing during the generation process. 
    It ensures that the questions are generated in a consistent format, facilitating easy integration into the rest of our pipeline.
    
    To allow qualified facts in \convkg, we generalized the standard triples to tuples with the additional relational predicate field.
    Note that while turn-specific objects are disallowed in the questions, objects from other turns that belong to the same predicate are encouraged to help create more complex questions.
For example, the query ``voice of [a] in [movie]'' could correspond to turn 2 of the prompt with the answer ``[b]''.

These configurations within the prompt are designed to maximize the effectiveness and applicability of the synthetic questions, making them fundamental to generating realistic and varied conversational QA instances. 
Accounting for linguistic variability and contextual appropriateness enables \convkg to curate robust, scalable, and highly configurable conversational KGQA datasets.
\begin{table*}[t]
  \centering
    \resizebox{0.7\textwidth}{!}{%
  \begin{tabular}{l|cc|cc|c}
    \toprule
    Dataset & \# Entities & \# Facts & \# Unique Types & \# Unique Predicates & \# Questions Per Fact   \\
    \midrule
    \emph{General} & 29M & 196M & 274 & 1252 & 24  \\
    \emph{Related} & 210K & 6.1M & 95 & 265 & 54  \\
    \bottomrule
  \end{tabular}}

  \caption{Dataset statistics for the various settings considered.}
  \label{tab:dataset_characteristics}
\end{table*}

\subsection{Conv.\ Factoid QA Instance Creation}
\label{subsec:convqa_filling}
Finally, a subset of extracted facts for an entity $e$, along with those for its related entities (if they exist), can be slot-filled using examples from the generated templates to get a conversation instance.
Note that this instance creation step adheres to some rules.
Regardless of the interaction type and selected linguistic phenomena, the first turn never involves any deixis.
We group certain predicates to ensure cohesiveness and avoid weird artifacts in the final conversations. 
For instance, questions about date of birth or place of birth are likely to occur near each other instead of being separated by several facts.

This process combines fresh, up-to-date factual data from the KG with synthetic templates, examinable by humans, to form a factoid KGQA instance.
Given that templatized generation and slot-filling are significantly cheaper than generating specific conversations for each new entity, \convkg allows us to curate large-scale, configurable datasets efficiently.
\section{Experimental Setup}
\label{sec:exp}

In all our experiments, we utilize a Wikidata dump with a knowledge cut-off date of June 2023.
To ensure consistency within our pipeline, which assumes English-language input, we filter the dump to only include entities with English names.
The original Wikidata dump contained approximately 100 million entities. However, after a meticulous cleaning pipeline and filtering to include only English examples and entities of \emph{interesting} types, our final dataset comprises 29 million entities associated with 196 million facts.

For the LLM predicate selector, we employed the \gptfour model, utilizing the \texttt{gpt-4-0613} endpoint provided by OpenAI. 
To maintain the model's reasoning efficacy and prevent cognitive overload, we limit our LLM requests to a maximum of 50 predicates at a time.
We implement this selection process in a segmented manner, ensuring comprehensive coverage of each type--predicate pair.
For predicates with linked qualifiers, we include the relationship predicate in the input to provide the necessary context.
This methodical approach allows us to isolate predicates that are particularly relevant for conversational factoid QA.

For the generation of synthetic question templates, we utilize the \texttt{gpt-3.5-turbo} endpoint.
We provide two in-context examples for each prompt (omitted for length) to better align generations to the expected template format.

For textual interactions, while calling the model endpoint, we leverage the ``logit\_bias'' field to penalize the model when it generates one of the question words --- wh-words or how.
Without doing this, we found the model ignores instructions and in-context examples to generate fully-formed questions instead.

For typo augmentation, going over each turn's question, we select at random one of the following attacks part of the \texttt{TextAttack} framework~\citep{morris2020textattack}: \texttt{WordSwapRandomCharacterDeletion()}, \texttt{WordSwapNeighboringCharacterSwap()}, or \texttt{WordSwapQWERTY()} and apply it.
We introduce a single ``meaningful'' typo to each question turn.

\begin{table*}[t]
  \centering
  \resizebox{0.92\textwidth}{!}{%
    \begin{tabular}{
      @{}cl|cccc|S[table-format=1.2]S[table-format=1.2]S[table-format=1.2]S[table-format=1.2]S[table-format=2.1, table-space-text-post={*}]@{}
    }
    \toprule
     & {Model} & {Interaction} & {Deixis} & {Disfluency} &  {Typo} & {Fluency} & {Relevance} & {Diversity} & {Grammar} & {Agreement} \\
    \midrule
    (1) & \convkgg & \voice & \xmark & \xmark & - & 3.97 & 4.63 & 2.40 & 3.90 & 75.5 \\
    (2) & \convkgg & \voice & \xmark & \cmark & - & 3.39 & 4.49 & 2.25 & 3.37 & 73.5 \\
    (3) & \convkgg & \voice & \cmark & \xmark & - & 3.99 & 4.59 & 2.45 & 3.77 & 74.8 \\
    (4) & \convkgg & \voice & \cmark & \cmark & - & 3.29 & 4.41 & 2.32 & 3.02 & 71.0 \\
    \midrule
    (5) & \convkgr & \voice & \xmark & \xmark & - & 3.70 & 3.71 & 2.66 & 3.69 & 68.5 \\
    (6) & \convkgr & \voice & \xmark & \cmark & - & 3.34 & 3.74 & 2.59 & 3.39 & 67.6 \\
    (7) & \convkgr & \voice & \cmark & \xmark & - & 3.76 & 3.89 & 2.79 & 3.72 & 69.3 \\
    (8) & \convkgr & \voice & \cmark & \cmark & - & 3.36 & 3.73 & 2.73 & 3.38 & 71.5 \\
    \midrule
    (9) & \convkgg & \search & \xmark & - & \xmark & 2.83 & 4.41 & 2.19 & 2.95 & 70.8 \\
    (10) & \convkgg & \search & \xmark & - & \cmark & 2.61 & 4.36 & 2.17 & 2.18 & 68.8 \\
    (11) & \convkgg & \search & \cmark & - & \xmark & 2.84 & 4.36 & 2.29 & 2.83 & 67.1 \\
    (12) & \convkgg & \search & \cmark & - & \cmark & 2.29 & 4.09 & 2.00 & 1.63 & 73.0  \\
        \midrule
    (13) & \convkgr & \search & \xmark & - & \xmark & 2.57 & 3.38 & 2.58 & 2.75 & 66.3 \\ 
    (14) & \convkgr & \search & \xmark & - & \cmark & 2.29 & 3.45 & 2.45 & 1.97 & 71.5 \\
    (15) & \convkgr & \search & \cmark & - & \xmark & 2.48 & 3.33 & 2.54 & 2.73 & 70.8 \\
    (16) & \convkgr & \search & \cmark & - & \cmark & 2.12 & 3.31 & 2.58 & 1.86 & 68.5 \\
    \bottomrule
    \end{tabular}
  }
    \caption{The results from the Single Model Rating of the \emph{General} (\convkgg) and \emph{Related} (\convkgr) set reflecting Likert scores of 1-5 for Fluency, Relevance, Diversity, and Grammar.
    Agreement scores represent the mean percentage of all scores where at least two of three annotators agree.}

  \label{tab:singresults}
\end{table*}
\section{Dataset Statistics}
\label{sec:data_stat}

\autoref{tab:dataset_characteristics} presents high-level statistics of the two large sets of data we curated using \convkg.

The \emph{General} set encompasses all entities and their associated facts from our filtered Wikidata, without incorporating any notion of related entities. 
This comprehensive collection comprises 29 million entities and an extensive 196 million facts.
For each fact, our methodology generates 24 possible questions: 12 from voice interactions (three each from original, deixis, disfluencies, and deixis\_disfluencies sets), and 12 from textual interactions (three each from original, deixis, typos, and deixis\_typos sets).
This approach exponentially increases the potential for generating diverse conversations, providing a large-scale resource for training conversational agents and exposing large language models to high-quality synthetic data.
The dataset's complexity and realism are further enhanced by its inclusion of 274 unique types and 1,252 unique predicates. 
This level of scale and coverage is challenging to achieve and is not typically observed in human-curated datasets.
For instance, \convquestions \cite{10.1145/3357384.3358016} contains only 11,200 real-user conversations, with an average of five questions each, derived from just five primary entity types.

The \emph{Related} set, while more specialized than the \emph{General} set, offers a focused exploration of popular Human-type entities, comprising 210,000 entities associated with 6.1 million facts. 
Despite its smaller scale, it provides a higher density of question variants, with an average of 54 questions per fact. 
This includes the 24 questions generated for the \emph{General} set, plus an additional 30 questions derived from related entity-specific follow-up queries, enabling a more detailed and nuanced exploration of topics. 
Encompassing 95 unique types and 265 unique predicates, this targeted dataset facilitates in-depth exploration and evaluation of conversational systems focused on human-centric entities.
The specialized structure of the \emph{Related} set makes it particularly valuable for assessing the effectiveness of AI systems in handling complex, interconnected queries about popular entities.

\section{Results}
\label{sec:results}
To evaluate the efficacy of \convkg, we employ three methods to comprehensively understand its quality and usefulness: (1) Single-Model Rating, (2) Pairwise Comparison, and (3) Parametric Knowledge Evaluation of LLMs.

Combining these methods aims to complement the other's strengths and weaknesses.
Likert scores, typically used in single-model grading, while very scalable as a human evaluation method, have several inherent limitations when used to evaluate language models. 
It relies on annotators making absolute judgments rather than relative comparisons, which tends to be less reliable for humans~\cite{Stewart2005-ne}.
The result can be inconsistent biases between different annotators~\cite{kulikov-etal-2019-importance}.
While pairwise comparisons avoid some of these issues by having annotators make relative judgments between pairs of data points, comparing a set of models is less efficient, often requiring re-evaluation of existing baseline models whenever a new model is introduced~\cite{Stewart2005-ne}.
Finally, we explore how LLMs fare at the synthetically generated conversational factoid QA datasets generated by \convkg, investigating their fact recall abilities by leveraging LLM-as-a-Judge evaluation and adding a critical dimension to our story. 
While this scales better while often correlating strongly with human annotations, they still suffer from issues like the self-enhancement bias, where LLM may favor the answers generated by themselves~\cite{llm-as-a-judge-ca}.
Together, these methods provide a robust and multifaceted approach to thoroughly evaluating the efficacy of \convkg, ensuring a comprehensive assessment from both human and automated perspectives.

\subsection{Single-Model Rating}
\label{smratin}
The Single-Model Rating task presents human annotators with a multi-turn conversation, in which they assign a score from 1-5 across four parameters.

We evaluated the dataset across four key parameters: Fluency, Relevance, Diversity, and Grammar. 
We assessed these parameters for 16 different combinations of settings available in the \convkg pipeline across 1600 conversations sampled with a uniform distribution, including Interaction (Voice or Text), Deixis (On or Off), Disfluency (On or Off, only for Voice), Typo (On or Off, only for Text), and Related Entities (On or Off).
At a high level, we designed it to cover a diverse set of unique entities sampled from Wikidata, featuring a wide range of entity types such as Person, Actor, Singer, and Politician.

The task interface, which we designed on an internal annotation tool, and in-depth guidelines are in \autoref{annotation}, including a detailed explanation of the crowdsourcing process (onboarding, training, and quality).

\autoref{tab:singresults} analyzes the alignment of the scores for each parameter with the 16 setting combinations.
The introduction of typographical errors affects fluency and grammar quality as their presence can disrupt the smooth flow and grammatical accuracy of the conversations.
Conversely, the inclusion of deixis can result in better fluency by creating more natural and contextually grounded conversations. 
However, deixis also impacts grammar, as referential expressions may introduce ambiguity or inconsistency in the dialogue structure.

As we would prefer, relevance and diversity appear resistant to variations in deixis, disfluencies, typographical errors, and interaction settings. 
The finding suggests that the content and informational diversity of the conversations remain largely unaffected by these factors. 
However, related entities impact the scores for relevance and diversity. 
By incorporating information from related entities, the conversations exhibit improved relevance to the topic and offer a wide range of knowledge discovery through the traversal of connected concepts in the KG.

Our analysis reveals that the optimal combination of settings for \convkg involves the voice interaction type with deixis and related entities. 
This configuration generates conversations that resemble natural human speech and discourse patterns, as reflected in the corresponding evaluation scores, minus disfluencies.
Despite the inherent subjectivity of human evaluation, the conversations generated by \convkg exhibit an average annotator agreement of 70.53\%, indicating a good level of consensus in their assessments and supporting the reliability of our evaluation.

These findings highlight the importance of considering multiple dimensions and interaction setting combinations when evaluating the quality of synthetically generated conversational datasets. 
By systematically exploring the impact of different factors on key evaluation parameters, we can gain a more nuanced understanding of the strengths and limitations of the \convkg approach. 
This knowledge can inform future pipeline refinements to enhance the quality and naturalness of the generated conversations.

Since the datasets curated by \convkg feature a diverse set of synthetically curated conversational QA instances and cover various entity types, linguistic phenomena, and interaction modalities, our benchmark can comprehensively evaluate a model's ability to handle the nuances and challenges of real-world conversations.

It is critical to note that single-model dialogue grading may be affected by a lack of relative understanding compared to other datasets and curation methods.

\begin{table}[t] 
  \centering
  \resizebox{\linewidth}{!}{%
    \begin{tabular}{
      @{}l
      S[table-format=2.1]
      S[table-format=2.1]
      S[table-format=2.1]
      S[table-format=2.1]@{}
    }
    \toprule
    Type & {Fluency (\%)} & {Relevance (\%)} & {Diversity (\%)} & {Grammar (\%)} \\
    \midrule
    Preference & 45.0 & 62.2 & 56.0 & 56.6 \\
    Agreement  & 84.6 & 89.0 & 82.2 & 86.6 \\
    \bottomrule
    \end{tabular}
  }
  \caption{The results from the Pairwise Comparison. We indicate pairwise comparisons through Preference, i.e., the percentage of graders who prefer \convkg.}
  \vspace{-1.6em}
  \label{tab:compresults}
\end{table}
\vspace{-0.6em}

\begin{table*}[t]
  \centering
  \resizebox{0.85\textwidth}{!}{
  \begin{tabular}{@{}cl|cccc|ccc@{}}
    \toprule
     & {Model} & {Interaction} & {Deixis} & {Disfluency} &  {Typo} & {Mean (\emph{Turn})} & {Mean (\emph{Conv.})} & {NA Ratio} \\
    \midrule
    (1) & GPT\(_{3.5}\) & \voice & \xmark & \xmark & - & 0.246 / 0.326 & 0.234 / 0.323 & 0.485 / 0.304 \\
    (2) & GPT\(_{3.5}\) & \voice & \xmark & \cmark & - & 0.250 / 0.349 & 0.236 / 0.346 & 0.434 / 0.272 \\
    (3) & GPT\(_{3.5}\) & \voice & \cmark & \xmark & - & 0.261 / 0.305 & 0.244 / 0.303 & 0.440 / 0.312 \\
    (4) & GPT\(_{3.5}\) & \voice & \cmark & \cmark & - & 0.261 / 0.306 & 0.254 / 0.304 & 0.432 / 0.276 \\
    \midrule
    (5) & GPT\(_{3.5}\) & \search & \xmark & - & \xmark & 0.246 / 0.333 & 0.233 / 0.329 & 0.459 / 0.276 \\
    (6) & GPT\(_{3.5}\) & \search & \xmark & - & \cmark & 0.220 / 0.279 & 0.199 / 0.277 & 0.513 / 0.352 \\
    (7) & GPT\(_{3.5}\) & \search & \cmark & - & \xmark & 0.239 / 0.307 & 0.221 / 0.302 & 0.445 / 0.306 \\
    (8) & GPT\(_{3.5}\) & \search & \cmark & - & \cmark & 0.201 / 0.220 & 0.179 / 0.219 & 0.519 / 0.433 \\
    \midrule
    \midrule
    (9) & GPT\(_4\) & \voice & \xmark & \xmark & - & 0.301 / 0.391 & 0.292 / 0.387 & 0.352 / 0.252 \\
    (10) & GPT\(_4\) & \voice & \xmark & \cmark & - & 0.320 / 0.412 & 0.307 / 0.407 & 0.329 / 0.232 \\
    (11) & GPT\(_4\) & \voice & \cmark & \xmark & - & 0.299 / 0.374 & 0.288 / 0.370 & 0.333 / 0.269 \\
    (12) & GPT\(_4\) & \voice & \cmark & \cmark & - & 0.299 / 0.384 & 0.290 / 0.381 & 0.340 / 0.244 \\
    \midrule
    (13) & GPT\(_4\) & \search & \xmark & - & \xmark & 0.316 / 0.371 & 0.294 / 0.366 & 0.335 / 0.285 \\
    (14) & GPT\(_4\) & \search & \xmark & - & \cmark & 0.265 / 0.347 & 0.242 / 0.346 & 0.451 / 0.350 \\
    (15) & GPT\(_4\) & \search & \cmark & - & \xmark & 0.269 / 0.361 & 0.248 / 0.355 & 0.385 / 0.309 \\
    (16) & GPT\(_4\) & \search & \cmark & - & \cmark & 0.222 / 0.290 & 0.201 / 0.285 & 0.479 / 0.396 \\
    \bottomrule
  \end{tabular}}
   \caption{The effectiveness based on the GPT\(_4\)-EVAL metric of two models GPT\(_{3.5}\) and GPT\(_{4}\) when evaluated against variants of the \emph{General} and \emph{Related} settings (scores separated by /). Note that all these settings are grounded on the same set of facts.}
   \vspace{-1.7em}
 \label{tab:merged_set}
\end{table*}
\subsection{Pairwise Comparison}
\label{pwcomp}

The Pairwise Comparison task introduces human annotators to two conversations: (1) a conversation generated by \convkg and (2) a commonly used conversational KGQA dataset. 
For these two conversations, the annotators indicate their preferences across the same psychometric evaluation metrics outlined in Section~\ref{smratin} across 500 conversations focusing on voice interaction, without disfluencies and related entities. 

The reference conversational QA dataset, \convquestions~\cite{10.1145/3357384.3358016}, was chosen based on its similarity to \convkg's purpose and capabilities while being human-curated.

The dataset generated with \convkg adapted the process outlined in Section~\ref{smratin} with three changes to mirror the attributes of the benchmark dataset: we restrict entity types to the ones included in the benchmark dataset, use the entity referred to in the first turn of the reference conversation as the starter entity in \convkg, and ensure the number of turns of both datasets is equal.

\convkg demonstrates varying qualitative preference across the psychometric schema compared to human-curated reference conversations as shown in Table \ref{tab:compresults}. 
In terms of fluency, \convkg nearly reaches parity with a preference of 45. 0\%. 
This slight underperformance can be attributed to the writing style generated by language models, which at times may not realistically represent natural writing patterns.
Additionally, inter-turn questions could lack the same cohesiveness as human-written conversations, given that we prioritize scale.

However, \convkg achieves significantly higher relevance with a 62.2\% preference. 
We believe this is due to the methodology employed in \convkg, which generates questions from a diverse knowledge base encompassing the primary and related entities. 
In contrast, human-curated conversations rely on annotators researching the given entity to create the dialogues, potentially leading to higher variability and divergence from the initial entity.

\convkg shows modest improvements in diversity and grammar, with preference rates of 56.0\% and 56.6\% respectively. 
The slight advantage in grammar may be due to the standardized dialect and writing style of the LLM utilized in \convkg, compared to the inherent variance across human annotators. 
The diversity improvement, while notable, still leaves room for enhancement. 
This limitation likely stems from the structured method of generating questions based on entity types and relationships in the KG, which could constrain the range of topics compared to the more open-ended human curation process.

It's worth noting that the agreement among human annotators is exceedingly high for all ratings, ranging from 82.2\% to 89.0\%. 
This strong consensus lends credibility to the evaluation results and suggests a clear differentiation between \convkg-generated and human-curated conversations across the assessed dimensions.

Overall, the human evaluation of \convkg reveals that it surpasses or closely reaches parity with human-curated conversations across four key dimensions: fluency, relevance, diversity, and grammar. 
These findings challenge the common perception that synthetically generated datasets are inherently of lower quality.
Instead, \convkg presents itself as a promising approach for generating high-quality conversational data in a repeatable and scalable manner. 

\subsection{Quantitative Analysis --- Parametric Knowledge Evaluation}
We explore the effectiveness of LLMs on 100 examples from each of the \emph{General} and \emph{Related} sets.
For each set, \convkg generates conversations spanning \emph{all} configurations considered.
This grounding on a consistent fact set enables us to test different hypotheses in a confounder-free manner, allowing us to carefully analyze how LLMs fare specifically with typos or a combination of deixis and disfluences in the voice interaction setup.

\autoref{fig:prompt.convqa} in \autoref{sec:add_prompts} presents an example interaction of how we evaluate LLMs on the conversational sets, iteratively as we go through each interaction turn of the conversational dataset.
For each turn, we prepend the model with the \emph{gold conversational history}.
The prompt is designed to ensure that the model provides the most accurate and relevant information directly about the query, omitting extraneous details.
In addition, the model can return responses in list form when multiple valid responses exist, ensuring clarity in the presentation of information.
Finally, the prompt allows the return of ``NA'' if low in confidence, i.e., if it believes the knowledge captured in its parameters or the ambiguity in the question results in it being unable to answer the question accurately.
The datasets were tested with two LLMs, \gptthreefive and \gptfour.

Upon curating factoid answers from these models, we employ \gptfour as a judge to rate the predictions in a binary fashion, as depicted in \autoref{fig:prompt.convqa_eval} in \autoref{sec:add_prompts}. 
Our evaluation prompt systematically assesses the correctness of responses.
Each candidate answer is compared against the gold answer for each conversational turn.
A score of 1 is assigned if the candidate addresses the query as per the gold standard; otherwise, a 0.
Finally, we instruct the model to score list answers with a score of 1 if some candidate string matches any gold answer.
The entire process respects the order of the conversation, providing scores in a list format that directly corresponds to the sequence of turns.

The results from this evaluation setup provide quantifiable metrics on the effectiveness of the tested LLMs, especially given that a metric like F1 and EM fails to accurately account for various aliases and other variations in LLM answers.

\autoref{tab:merged_set} presents the \gptfour-EVAL results for the variants from the \emph{General} and \emph{Related} settings.
The metrics include the mean score assigned at a turn or conversation level.
Additionally, we also report the rate of refusals (NA Ratio).

Firstly, we note that in both tables, \gptfour scores higher across the board than \gptthreefive, rows (9)--(16) vs.\ (1)--(8).
This finding could be due to the enhanced capabilities of the larger models, which also benefit from more extensive training data and refined instruction fine-tuning.
The drop in the refusals is also indicative of \gptfour having successfully stuffed a lot more information in the model parameters than the smaller \gptthreefive.

Second, we can see that it is inconclusive whether conversations in the voice interaction setting achieve higher scores than those in the textual interaction setting when not compounded by other linguistic phenomena, as illustrated in rows (1) vs. (5) and (9) vs. (13).
In the presence of deixis, the outcomes are similarly nuanced, although slightly favoring the voice interaction setting, rows (3) vs.\ (7) and (11) vs.\ (15).
This result suggests that these models can more easily resolve referents in spoken queries, which often contain more contextual clues than text-based keyword queries.

Third, the introduction of disfluencies, unexpectedly, appears to have a negligible or even slightly beneficial effect on the results in voice interaction settings, rows (2) vs.\ (1), (4) vs.\ (3), (10) vs.\ (9), and (12) vs.\ (11).
These findings indicate that LLMs are becoming increasingly adept at filtering out irrelevant signals to focus on the core informational need of a query.

Finally, typos, as one might predict, diminish the effectiveness of both models. 
This is reflected in the decrease in scores whenever typos are present, rows (6) vs.\ (5), (8) vs.\ (4), (14) vs.\ (13), and (16) vs.\ (15), underscoring the models' sensitivity to correct spelling as a factor in understanding and processing questions.

Overall, these results provide a nuanced understanding of LLMs in the domain of conversational factoid question answering across diverse configurable settings.  
We posit that a comprehensive evaluation encompassing this array of configurations is imperative to develop a thorough portrayal of system effectiveness.

\section{Conclusions}
\label{sec:conclusions}

In this paper, we introduce \convkg, a novel framework for generating dynamic and scalable conversational datasets for Knowledge Graph Question Answering (KGQA). Our system leverages the structured representation of Knowledge Graphs (KGs) to produce configurable and adaptive conversational datasets that evolve with user information needs and KG-captured knowledge.
Extensive evaluations demonstrate \convkg's effectiveness in generating very high-quality KGQA datasets. 
Through rigorous qualitative and quantitative tests, we showcase the versatility of these datasets across various conversational scenarios, enabling the assessment of models' effectiveness in different facets of user interactions and linguistic phenomena.

\convkg enhances the testing capabilities of LLMs and QA systems in adapting to the ever-growing knowledge landscape but also facilitates high-quality evaluation across different forms of user interactions, each with its nuances.

\section*{Acknowledgement}

We would like to thank Anil Pacaci, Simone Conia and Varun Embar for insightful discussions surrounding the choices made during the project. 
\bibliography{main}

\begin{thebibliography}{25}
\providecommand{\natexlab}[1]{#1}

\bibitem[{Barba et~al.(2021)Barba, Pasini, and Navigli}]{barba-etal-2021-esc}
Edoardo Barba, Tommaso Pasini, and Roberto Navigli. 2021.
\newblock {ESC}: Redesigning {WSD} with extractive sense comprehension.
\newblock In \emph{Proceedings of the 2021 Conference of the North American Chapter of the Association for Computational Linguistics: Human Language Technologies}. Association for Computational Linguistics.

\bibitem[{Chakrabarti et~al.(2022)Chakrabarti, Singh, Lohiya, Jain, and ~}]{chakrabarti-etal-2022-joint}
Soumen Chakrabarti, Harkanwar Singh, Shubham Lohiya, Prachi Jain, and Mausam ~. 2022.
\newblock Joint completion and alignment of multilingual knowledge graphs.
\newblock In \emph{Proceedings of the 2022 Conference on Empirical Methods in Natural Language Processing}. Association for Computational Linguistics.

\bibitem[{Christmann et~al.(2023)Christmann, Roy, and Weikum}]{christmann2023compmix}
Philipp Christmann, Rishiraj~Saha Roy, and Gerhard Weikum. 2023.
\newblock Compmix: A benchmark for heterogeneous question answering.
\newblock \emph{arXiv:2306.12235}.

\bibitem[{Christmann et~al.(2019)Christmann, Saha~Roy, Abujabal, Singh, and Weikum}]{10.1145/3357384.3358016}
Philipp Christmann, Rishiraj Saha~Roy, Abdalghani Abujabal, Jyotsna Singh, and Gerhard Weikum. 2019.
\newblock Look before you hop: Conversational question answering over knowledge graphs using judicious context expansion.
\newblock In \emph{Proceedings of the 28th ACM International Conference on Information and Knowledge Management}, CIKM '19. Association for Computing Machinery.

\bibitem[{Conia et~al.(2023)Conia, Li, Lee, Minhas, Ilyas, and Li}]{conia-etal-2023-increasing}
Simone Conia, Min Li, Daniel Lee, Umar Minhas, Ihab Ilyas, and Yunyao Li. 2023.
\newblock Increasing coverage and precision of textual information in multilingual knowledge graphs.
\newblock In \emph{Proceedings of the 2023 Conference on Empirical Methods in Natural Language Processing}. Association for Computational Linguistics.

\bibitem[{De~Cao et~al.(2022)De~Cao, Wu, Popat, Artetxe, Goyal, Plekhanov, Zettlemoyer, Cancedda, Riedel, and Petroni}]{de-cao-etal-2022-multilingual}
Nicola De~Cao, Ledell Wu, Kashyap Popat, Mikel Artetxe, Naman Goyal, Mikhail Plekhanov, Luke Zettlemoyer, Nicola Cancedda, Sebastian Riedel, and Fabio Petroni. 2022.
\newblock Multilingual autoregressive entity linking.
\newblock \emph{Transactions of the Association for Computational Linguistics}.

\bibitem[{Guo et~al.(2022)Guo, Zhuang, Qin, Zhu, Xie, Xiong, and He}]{guo-etal-2022-recommender}
Qingyu Guo, Fuzhen Zhuang, Chuan Qin, Hengshu Zhu, Xing Xie, Hui Xiong, and Qing He. 2022.
\newblock A survey on knowledge graph-based recommender systems.
\newblock \emph{IEEE Transactions on Knowledge and Data Engineering}.

\bibitem[{Guu et~al.(2020)Guu, Lee, Tung, Pasupat, and Chang}]{guu-etal-2020-realm}
Kelvin Guu, Kenton Lee, Zora Tung, Panupong Pasupat, and Mingwei Chang. 2020.
\newblock Retrieval augmented language model pre-training.
\newblock In \emph{Proceedings of the 37th International Conference on Machine Learning}, Proceedings of Machine Learning Research. PMLR.

\bibitem[{Hogan et~al.(2021)Hogan, Blomqvist, Cochez, D’amato, Melo, Gutierrez, Kirrane, Gayo, Navigli, Neumaier, Ngomo, Polleres, Rashid, Rula, Schmelzeisen, Sequeda, Staab, and Zimmermann}]{hogan-etal-2021-knowledge}
Aidan Hogan, Eva Blomqvist, Michael Cochez, Claudia D’amato, Gerard~De Melo, Claudio Gutierrez, Sabrina Kirrane, Jos\'{e} Emilio~Labra Gayo, Roberto Navigli, Sebastian Neumaier, Axel-Cyrille~Ngonga Ngomo, Axel Polleres, Sabbir~M. Rashid, Anisa Rula, Lukas Schmelzeisen, Juan Sequeda, Steffen Staab, and Antoine Zimmermann. 2021.
\newblock Knowledge graphs.
\newblock \emph{ACM Comput. Surv.}

\bibitem[{Kulikov et~al.(2019)Kulikov, Miller, Cho, and Weston}]{kulikov-etal-2019-importance}
Ilia Kulikov, Alexander Miller, Kyunghyun Cho, and Jason Weston. 2019.
\newblock Importance of search and evaluation strategies in neural dialogue modeling.
\newblock In \emph{Proceedings of the 12th International Conference on Natural Language Generation}. Association for Computational Linguistics.

\bibitem[{Ma et~al.(2022)Ma, Pradeep, Nogueira, and Lin}]{unicoil}
Xueguang Ma, Ronak Pradeep, Rodrigo Nogueira, and Jimmy Lin. 2022.
\newblock Document expansions and learned sparse lexical representations for ms marco {V1} and {V2}.
\newblock In \emph{Proceedings of the 45th Annual International ACM SIGIR Conference on Research and Development in Information Retrieval (SIGIR 2022)}.

\bibitem[{Morris et~al.(2020)Morris, Lifland, Yoo, Grigsby, Jin, and Qi}]{morris2020textattack}
John Morris, Eli Lifland, Jin~Yong Yoo, Jake Grigsby, Di~Jin, and Yanjun Qi. 2020.
\newblock Textattack: A framework for adversarial attacks, data augmentation, and adversarial training in nlp.
\newblock In \emph{Proceedings of the 2020 Conference on Empirical Methods in Natural Language Processing: System Demonstrations}.

\bibitem[{Nogueira and Lin(2019)}]{doct5query}
Rodrigo Nogueira and Jimmy Lin. 2019.
\newblock From doc2query to doctttttquery.

\bibitem[{Peng et~al.(2023)Peng, Galley, He, Cheng, Xie, Hu, Huang, Liden, Yu, Chen, and Gao}]{peng-etal-2023-check}
Baolin Peng, Michel Galley, Pengcheng He, Hao Cheng, Yujia Xie, Yu~Hu, Qiuyuan Huang, Lars Liden, Zhou Yu, Weizhu Chen, and Jianfeng Gao. 2023.
\newblock Check your facts and try again: Improving large language models with external knowledge and automated feedback.
\newblock \emph{arXiv:2302.12813}.

\bibitem[{Petroni et~al.(2019)Petroni, Rockt{\"a}schel, Riedel, Lewis, Bakhtin, Wu, and Miller}]{petroni-etal-2019-language}
Fabio Petroni, Tim Rockt{\"a}schel, Sebastian Riedel, Patrick Lewis, Anton Bakhtin, Yuxiang Wu, and Alexander Miller. 2019.
\newblock Language models as knowledge bases?
\newblock In \emph{Proceedings of the 2019 Conference on Empirical Methods in Natural Language Processing and the 9th International Joint Conference on Natural Language Processing (EMNLP-IJCNLP)}. Association for Computational Linguistics.

\bibitem[{Pradeep et~al.(2022)Pradeep, Li, Wang, and Lin}]{NQS}
Ronak Pradeep, Yilin Li, Yuetong Wang, and Jimmy Lin. 2022.
\newblock Neural query synthesis and domain-specific ranking templates for multi-stage clinical trial matching.
\newblock In \emph{Proceedings of the 45th International ACM SIGIR Conference on Research and Development in Information Retrieval}, SIGIR '22. Association for Computing Machinery.

\bibitem[{Pradeep et~al.(2023{\natexlab{a}})Pradeep, Sharifymoghaddam, and Lin}]{rankvicuna}
Ronak Pradeep, Sahel Sharifymoghaddam, and Jimmy Lin. 2023{\natexlab{a}}.
\newblock {RankVicuna}: Zero-shot listwise document reranking with open-source large language models.
\newblock \emph{arXiv:2309.15088}.

\bibitem[{Pradeep et~al.(2023{\natexlab{b}})Pradeep, Sharifymoghaddam, and Lin}]{rankzephyr}
Ronak Pradeep, Sahel Sharifymoghaddam, and Jimmy Lin. 2023{\natexlab{b}}.
\newblock {RankZephyr}: Effective and robust zero-shot listwise reranking is a breeze!
\newblock \emph{arXiv:2312.02724}.

\bibitem[{Reinanda et~al.(2020)Reinanda, Meij, and de~Rijke}]{reinanda-etal-2020-knowledge}
Ridho Reinanda, Edgar Meij, and Maarte de~Rijke. 2020.
\newblock Knowledge graphs: An information retrieval perspective.
\newblock \emph{Foundations and Trends{\textregistered} in Information Retrieval}.

\bibitem[{Schneider et~al.(2022)Schneider, Schopf, Vladika, Galkin, Simperl, and Matthes}]{schneider-etal-2022-decade}
Phillip Schneider, Tim Schopf, Juraj Vladika, Mikhail Galkin, Elena Simperl, and Florian Matthes. 2022.
\newblock A decade of knowledge graphs in natural language processing: A survey.
\newblock In \emph{Proceedings of the 2nd Conference of the Asia-Pacific Chapter of the Association for Computational Linguistics and the 12th International Joint Conference on Natural Language Processing (Volume 1: Long Papers)}. Association for Computational Linguistics.

\bibitem[{Stewart et~al.(2005)Stewart, Brown, and Chater}]{Stewart2005-ne}
Neil Stewart, Gordon D~A Brown, and Nick Chater. 2005.
\newblock Absolute identification by relative judgment.
\newblock \emph{Psychol. Rev.}

\bibitem[{Tamber et~al.(2023)Tamber, Pradeep, and Lin}]{lit5}
Manveer~Singh Tamber, Ronak Pradeep, and Jimmy Lin. 2023.
\newblock Scaling down, litting up: Efficient zero-shot listwise reranking with seq2seq encoder-decoder models.
\newblock \emph{arXiv:2312.16098}.

\bibitem[{Xian et~al.(2024)Xian, Samuel, Khoubsirat, Pradeep, Sultan, Florian, Roukos, Sil, Potts, and Khattab}]{path}
Jasper Xian, Saron Samuel, Faraz Khoubsirat, Ronak Pradeep, Md~Arafat Sultan, Radu Florian, Salim Roukos, Avirup Sil, Christopher Potts, and Omar Khattab. 2024.
\newblock {Prompts as Auto-Optimized Training Hyperparameters}: Training best-in-class {IR} models from scratch with 10 gold labels.
\newblock \emph{arXiv:2406.11706}.

\bibitem[{Xu et~al.(2023)Xu, Namazifar, Hazarika, Padmakumar, Liu, and Hakkani-T{\"u}r}]{xu-etal-2023-kilm}
Yan Xu, Mahdi Namazifar, Devamanyu Hazarika, Aishwarya Padmakumar, Yang Liu, and Dilek Hakkani-T{\"u}r. 2023.
\newblock {KILM}: Knowledge injection into encoder-decoder language models.
\newblock \emph{arXiv:2302.09170}.

\bibitem[{Zheng et~al.(2023)Zheng, Chiang, Sheng, Zhuang, Wu, Zhuang, Lin, Li, Li, Xing, Zhang, Gonzalez, and Stoica}]{llm-as-a-judge-ca}
Lianmin Zheng, Wei-Lin Chiang, Ying Sheng, Siyuan Zhuang, Zhanghao Wu, Yonghao Zhuang, Zi~Lin, Zhuohan Li, Dacheng Li, Eric.~P Xing, Hao Zhang, Joseph~E. Gonzalez, and Ion Stoica. 2023.
\newblock Judging llm-as-a-judge with mt-bench and chatbot arena.
\newblock \emph{arXiv:2306.05685}.

\end{thebibliography}

\clearpage
\newpage

\appendix

\begin{figure*}[!ht]
\begin{mdframed}[roundcorner=10pt, linecolor=blue, linewidth=2pt, innerleftmargin=10pt, innerrightmargin=10pt, innertopmargin=10pt, innerbottommargin=10pt]
\textbf{SYSTEM:} You are a helpful assistant that can help select all predicates likely to be used in a Factoid Conversational QA dataset for a particular type of entity. You should not select something like id/index/phone number/Commons category (which does not lend well to Conversational QA), name (which is obvious from the question itself), and also things which have little or nothing to do with the particular type like goals scored for a type actor or supported sports team for a singer. Predicates whose corresponding objects have type video, audio, and image should also not be included.  Do not include first name and last name which would already be obvious from the user question. Things like marriage/partners should be included. You will be provided with a type and a table of tuples of the form (predicate\_id, predicate\_name). Always provide only an answer and in the format \"<pythonic list of useful predicate ids>\".

\textbf{USER:}
Type: \emph{singer}\\
Predicates: [('P412', 'voice type'),  ('P4431', 'Google Doodle'), ('P793', 'significant event'),  \ldots] 

\textbf{\gptfour:} \emph{[('P412', 'voice type'), \ldots]}
\end{mdframed}
\caption{Prompt for the LLM-based Predicate Selector.}
\label{fig:prompt.pfilter}
\end{figure*}
\begin{figure*}[!t]
\begin{mdframed}[roundcorner=10pt, linecolor=blue, linewidth=2pt, innerleftmargin=10pt, innerrightmargin=10pt, innertopmargin=10pt, innerbottommargin=4pt]
\textbf{SYSTEM:} You are an AI assistant tasked with generating a natural conversational question-answering session between two people, A and B, based on information from a knowledge graph, in the form of a list of triples.
A will only ask questions, and they should be based on the subject type and predicate of each triple, while B will only answer with just the object and no extraneous information.
To make the conversation more realistic, you should also include for A:\\
- deixis (words that refer to people, places, or things in the conversation history like this, their, that, it, they, them)\\
- disfluencies (pauses, repetitions, and other speech errors that occur naturally in conversation)\\
- deixis\_disfluencies (each question displays both deixis and disfluencies)\\
You only return JSON of the following form with key being an <int representing the turn number> mapping to:

- original: <list of three variants of standard single-turn questions not depending on conversation history answered by the answer field>\\
- deixis: <deixis applied to original variants>\\
- disfluencies: <disfluencies applied to original variants>\\
- deixis\_disfluencies: <disfluencies applied to deixis variants>\\
- answer: <always the object field from the turn triple, representing B's answer to any of the questions>\\
Ensure that the variants of the original have the subject variable (enclosed by []) as is and that the object is always the answer and is never part of the questions.
Ensure there are exactly three variants of each type.
All questions should mimic real world conversational questions.\\
\textbf{USER:}
You have been provided with K triples (subject, predicate, object) from the knowledge graph corresponding directly to exact turns.
The subject and object, in this case, are templates and enclosed by [], and the subject template should be used as is for questions in the original field.
For example, for a triple ([person], gender, [x]), a question in the original field should always use the literal "[person]" without any deixis.
The answer field should always be the turn's object template.
Your task is to use this information to generate a coherent conversational question-answering session between A and B following the aforementioned template.
Remember their roles exactly and ensure the conversation length is equal to the number of turns.

Examples:

\# Triples\\
Turn 1: ([cricketer], number of matches played/races/starts, [a]) \\
\vdots \\
\end{mdframed}
\caption{The prompt used for Synthetic Question Template Generation in the \emph{Voice} setting.}
\label{fig:prompt.question}
\end{figure*}

\begin{figure*}[!ht]
\begin{mdframed}[roundcorner=10pt, linecolor=blue, linewidth=2pt, innerleftmargin=10pt, innerrightmargin=10pt, innertopmargin=10pt, innerbottommargin=10pt]
\textbf{SYSTEM:} You are an AI assistant tasked with generating a natural conversational
question-answering session between two people, A and B, based on information 
from a knowledge graph, in the form of a list of triples. A will only ask 
questions, and they should be based on the subject type and predicate of each 
triple, while B will only answer with just the object and no extraneous 
information. To make the conversation more realistic, you should also include 
for A:\\
- deixis (words that refer to people, places, or things in the conversation history like this, their, that, it, they, them)\\
You only return JSON of the following form with key being an <int representing 
the turn number> mapping to:\\
- original: <list of three variants of standard single-turn questions not 
depending on conversation history answered by the answer field>\\
- deixis: <deixis applied to original variants>\\
- answer: <always the object field from the turn triple, representing B's answer
to any of the questions>\\
Ensure that the variants of the original have the subject variable (enclosed by 
[]) as is and that the object is always the answer and is never part of the
questions. Ensure there are exactly three variants of each type. All questions 
should mimic real world user search queries and be short, lower case and never
proper questions beginning with who/whom/what/when/which/how. Ensure to never 
generate proper questions for any variant of the four types of queries.

\textbf{USER}: You have been provided with K triples (subject, predicate, object) from the
knowledge graph corresponding directly to exact turns. The subject and object,
in this case, are templates and enclosed by [], and the subject template should
be used as is for questions in the original field. For example, for a triple
([person], gender, [x]), a question in the original field should always use the
literal "[person]" without any deixis. The answer field should always be the 
turn's object template. Your task is to use this information to generate a
coherent conversational question-answering session between A and B following the
aforementioned template. Remember their roles exactly and ensure the 
conversation length is equal to the number of turns.

Examples:
We see in the following examples all variants take on user search query form and
never start with one of a who, what, when, which, and how.\\
\# Triples\\
Turn 1: ([cricketer], number of matches played/races/starts, [a]) \\
Turn 2: ([cricketer], date of birth, [b]) \\
\vdots \\
\end{mdframed}
\caption{Prompt used for Synthetic Question Template Generation in the \emph{Text} (Search) setting.}
\label{fig:prompt.query}
\end{figure*}

\section{Additional Prompts}
\label{sec:add_prompts}
\begin{figure*}[!ht]
\begin{mdframed}[roundcorner=10pt, linecolor=blue, linewidth=2pt, innerleftmargin=10pt, innerrightmargin=10pt, innertopmargin=10pt, innerbottommargin=10pt]
\textbf{SYSTEM:} You are an AI assistant tasked with generating a natural conversational question-answering session between two people, A and B, based on information from a knowledge graph, in the form of a list of tuples.
A will only ask questions, and they should be based on the subject type, predicate, and relationship predicate of each tuple (potentially also an object from another tuple provided), while B will only answer with just the object and no extraneous information.
To make the conversation more realistic, you should also include for A:\\
- deixis (words that refer to people, places, or things in the conversation history like this, their, that, it, they, them) applied to just the subject template (never to any of the objects included)\\
You only return JSON of the following form with key being an <int representing the turn number> mapping to:

- original: <list of three variants of standard single-turn questions not depending on conversation history answered by the answer field>\\
- deixis: <deixis applied to original variants>\\
- answer: <always the object field from the turn tuple, representing B's answer to any of the questions>

Ensure that the variants of the original have the subject variable (enclosed by []) as is and that the object is always the answer and is never part of the questions.
Ensure there are exactly three variants of each type.
All questions should mimic real world user search queries and be short, lower case and never proper questions beginning with who/whom/what/when/which/how.
Ensure to never generate proper questions for any variant of the four types of queries.

\textbf{USER}: You have been provided with K tuples (subject, predicate, relationship\_predicate, object) from the knowledge graph corresponding directly to exact turns.
The subject and object, in this case, are templates and enclosed by [], and the subject template should be used as is for questions in the original field.
For example, for a tuple ([person], marriage, related person, [a]), a question in the original field should always use the literal "[person]" without any deixis.
You can also use the object field from any of the other tuples from the same predicate, if available, to craft better questions.
The answer field should always be the turn's object template.
Your task is to use this information to generate a coherent conversational question-answering session between A and B following the aforementioned template.
Remember their roles exactly and ensure the conversation length is equal to the number of turns.
Never use the object template corresponding to the turn ([a] in 1, [b] in 2, ...) in any of the turn's questions.

Examples:
We see in the following examples all variants take on user search query form and never start with one of a who, what, when, which, and how.

\# Triples\\
Turn 1: ([movie], voice actor, performer, [a])

Turn 2: ([movie], voice actor, character, [b])

\vdots
\end{mdframed}
\caption{Prompt used for Synthetic Question Template Generation in the \emph{Text} setting with Relationship Predicates.}
\label{fig:prompt.complex_query}
\end{figure*}

\begin{figure*}[!ht]
\begin{mdframed}[roundcorner=10pt, linecolor=blue, linewidth=2pt, innerleftmargin=10pt, innerrightmargin=10pt, innertopmargin=10pt, innerbottommargin=10pt]
\textbf{SYSTEM:} You are a helpful assistant that can do conversational factoid question answering. You only provide the exact answer span and never with extraneous information or in full sentences. Provide the answer in a string or pythonic list (the list can have multiple elements if there are multiple answers). Always provide an answer in the format "Answer: <answer string or list of answer strings>". If you are extremely unsure of the answer, return "Answer: NA".

\textbf{USER:} Who narrated the Penguins documentary?

\textbf{\gptfour:} \emph{Ed Helms}

\textbf{USER:} Ummm, who was, hmm, its director?

\textbf{\gptfour:}  \emph{Alastair Fothergill}

\end{mdframed}
\caption{Example Interaction for GPT\textsubscript{x} baselines of ConvKGYarn.}
\label{fig:prompt.convqa}
\end{figure*}
\begin{figure*}[!ht]
\begin{mdframed}[roundcorner=10pt, linecolor=blue, linewidth=2pt, innerleftmargin=10pt, innerrightmargin=10pt, innertopmargin=10pt, innerbottommargin=10pt]
\textbf{SYSTEM:} You are a helpful assistant that can help evaluate conversational factoid question answering. You will be provided Questions, Gold Answers, and Candidates, turn-by-turn. The Gold Answer and Candidate are either a single answer or list of answers.  If the Candidate seems to properly answer the question based on the answers, score it a 1, else, a 0. Do not use any of your global knowledge. If they are lists, ensure that at least one of the Candidate is captured by the Gold Answers. Do not use any additional knowledge. The output should be of the form \"Ratings: <pythonic list of 0s/1s>\" where the list\'s order corresponds exactly to the conversation turn

\textbf{USER:} Question: Who narrated the Penguins documentary?

Gold Answers: Ed Helms 
Candidates: Ed Helms

Question: Ummm, who was, hmm, its director? 

Gold Answers: Alastair Fothergill 
Candidates: NA

Question: Who produced the documentary? 

Gold Answers: [Alastair Fothergill, Keith Scholey, Roy Conli] 

Candidates: Scholey

\textbf{\gptfour:} \emph{[1, 0, 1]}

\vdots

\end{mdframed}
\caption{Prompt for \gptfour-eval  of ConvKGYarn.}
\label{fig:prompt.convqa_eval}
\end{figure*}
In this section, we include a few prompts that we could not include in Section~\ref{sec:method} because of space restrictions.
\autoref{fig:prompt.pfilter} illustrates the prompt for predicate filtering.
For simple and complex facts, the detailed prompt for generating templatized questions for voice and textual (or search) interactions are in \autoref{fig:prompt.question} and \autoref{fig:prompt.query}, respectively. 
For qualified facts, we provide the prompt used in \autoref{fig:prompt.complex_query}.

\autoref{fig:prompt.convqa} presents an example interaction of how we evaluate LLMs on the conversational sets, iteratively as we go through each interaction turn of the conversational dataset.
Upon curating factoid answers from these models, we employ GPT\textsubscript{4} as a judge to rate the predictions in a binary fashion, as depicted in \autoref{fig:prompt.convqa_eval}.

\section{Human Annotation Process}
\label{annotation}
In this section, we provide in-depth details on \convkg's human annotation process used during the evaluation tasks. The human annotation interface and associated guidelines are attached in the appendix.

\begin{figure*}[t]
    \centering
    \includegraphics[width=0.8\linewidth]{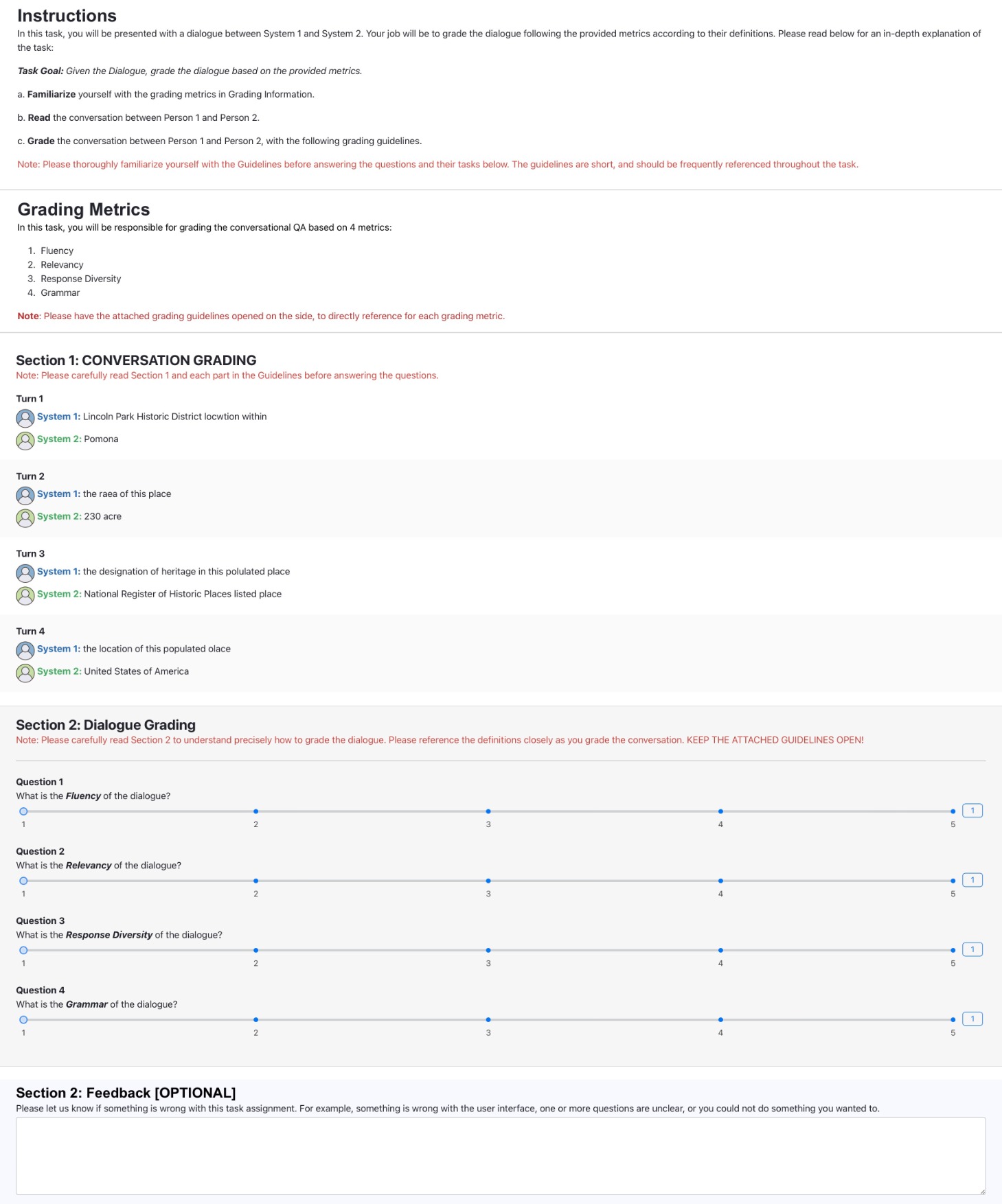}  
    \caption{The human annotation user interface for the Psychometric Evaluation of ConvKGYarn.}
    \label{fig:Grading_UI}
\end{figure*}

\subsection{Psychometric Evaluation}

The objective of the annotation process was to grade the provided conversation on a Likert scale of 5, across a defined psychometric evaluation schema.
First, given a conversation, the human annotators were asked to familiarize themselves with its information: the user interface for the task provided a short overview of the instructions, as well as the evaluation schema upon which the conversation would be graded. 
In addition, the annotators were provided with a thorough instruction file, which correlated directly to the annotation task and gave granular details on the task, the evaluation schema, and helpful tips.

After learning about the task, the annotators were tasked with grading the conversation across the provided evaluation schema on a scale of 1 to 5. To do so, human annotators were recommended to become thoroughly familiar with the context of the conversation.
The evaluation schema consisted of several psychometric dimensions, each with its own set of criteria and definitions. For each dimension, annotators could choose one of the following general options.
However, the definition and scaling explanation was tailored to each dimension, to provide a granular understanding.

\begin{itemize}
    \item 1 - Poor. The conversation fails to meet the criteria for the given dimension and exhibits significant issues or deficiencies.
    \item 2 - Fair. The conversation partially meets the criteria for the given dimension but has some notable weaknesses or areas for improvement.
    \item 3 - Satisfactory. The conversation adequately meets the criteria for the given dimension, with no major strengths or weaknesses.
    \item 4 - Good. The conversation effectively meets the criteria for the given dimension and demonstrates some notable strengths or positive qualities.
    \item 5 - Excellent. The conversation fully meets or exceeds the criteria for the given dimension, exhibiting exceptional quality or performance.
\end{itemize}

Annotators were given the choice to opt out from rating a conversation if they felt they did not have enough context or knowledge about the topic to make an informed assessment.

Please refer to the Dialogue Grading - Task Guidelines for further information on the evaluation schema and their definitions.

\begin{figure*}[t]
    \centering
    \includegraphics[width=0.8\linewidth]{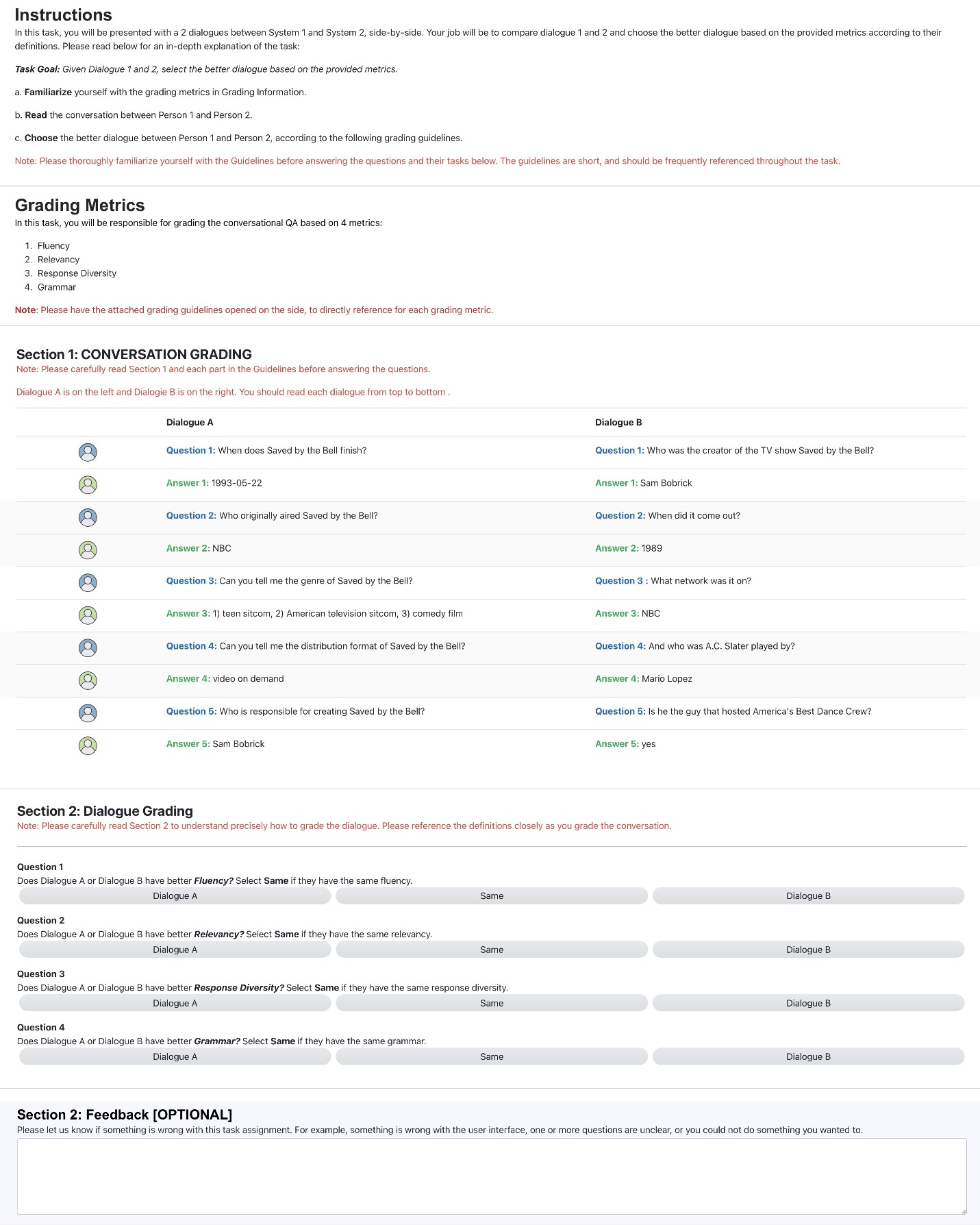}  
    \caption{The human annotation user interface for the Psychometric Comparative Analysis of ConvKGYarn.}
    \label{fig:Comparative_UI}
\end{figure*}

\subsection{Comparative Analysis}
Similar to the previous annotation task, the objective of this annotation process was to compare two conversations with a similar context, under the same psychometric evaluation schema.
The task undertaken by the human annotators was the main difference between the two annotation processes.

First, given a pair of conversations, the human annotators were asked to familiarize themselves with the information provided: the user interface for the task presented a short overview of the instructions, as well as the evaluation schema upon which the conversations would be compared.
In addition, the annotators were provided with a thorough instruction file, which correlated directly to the annotation task and gave granular details on the task, the evaluation schema, and helpful tips.

After learning about the task, the annotators were tasked with comparing the two conversations across the provided evaluation schema.
The evaluation schema consisted of several psychometric dimensions, each with its own set of criteria and definitions. For each dimension, annotators could choose one of the following options:

\begin{itemize}
\item Conversation A. The first conversation better meets the criteria for the given dimension compared to the second conversation.
\item Conversation B. The second conversation better meets the criteria for the given dimension compared to the first conversation.
\item Same. Both conversations equally meet the criteria for the given dimension, with no significant differences between them.
\end{itemize}

Please refer to the Dialogue Comparisons - Task Guidelines for further information on the evaluation schema and their definitions.

\subsection{Quality Assurance and Inter-Annotator Agreement}
Closely adapted from \citet{conia-etal-2023-increasing}, to ensure the highest quality output, all human annotators were required to pass a rigorous entrance test before participating in the annotation process. 
This test involved studying a comprehensive set of guidelines that familiarized the annotator with the fundamental concepts of conversational KGQA, outlined the task and UI elements, and provided illustrative examples. Additionally, annotators had to successfully complete qualification exams tailored to each specific task, achieving a pre-defined threshold compared to the gold labels. 
Only annotators who passed the entrance test were permitted to proceed with the actual annotation process (the 25 conversations used in the entrance test were excluded from the final dataset).

We exclusively recruited annotators who could demonstrate proficiency in English, and limited the locales to either en-US or en-CA. 
Compensation for annotators was based on the competitive hourly wages per annotator's geographic location. 
On average, annotators dedicated approximately 5 minutes to each conversation. 
Given that each conversation was evaluated by 3 annotators, we estimate the total human time invested in the annotation process to be 3 annotators $\times$ 1,000 conversations $\times$ 5 minutes / 60 minutes = 250 hours.

Upon completion of the annotation process, we assessed inter-annotator agreement using a majority vote calculation. 
Table 4 illustrates an average agreement of 70.53\% (Psychometric Evaluation) and 85.6\% (Comparative Analysis)  which is generally considered to be a strong level of agreement.

This inter-annotator agreement score serves to validate the results obtained from the annotation process.

\clearpage
\includepdf[pages=-]{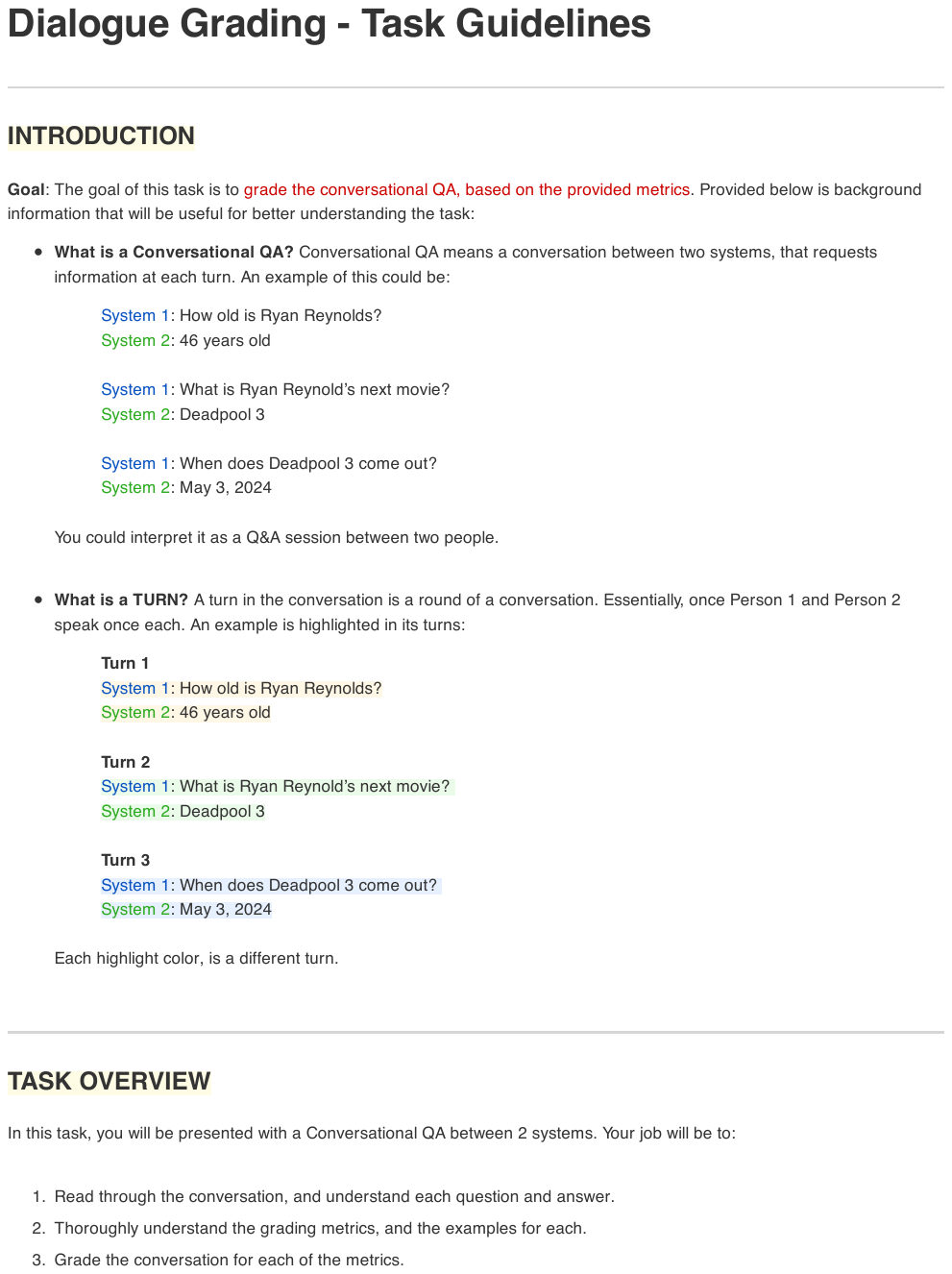}
\clearpage
\includepdf[pages=-]{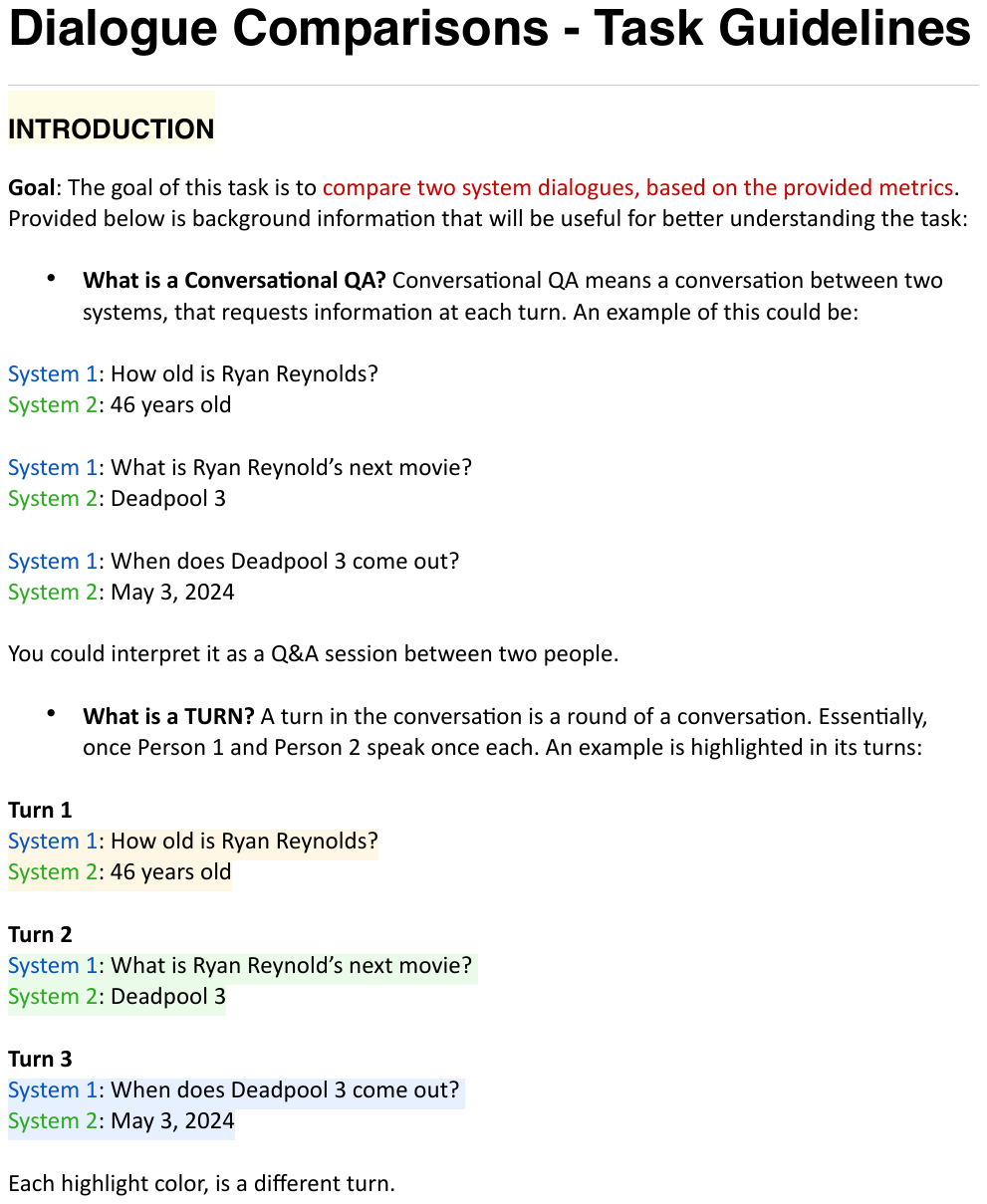}

\end{document}